\documentclass{article} 
\usepackage{preprint_style,times}

\usepackage{amsmath,amsfonts,bm}

\def\eqref#1{equation~\ref{#1}}

\def\1{\bm{1}}

\DeclareMathAlphabet{\mathsfit}{\encodingdefault}{\sfdefault}{m}{sl}
\SetMathAlphabet{\mathsfit}{bold}{\encodingdefault}{\sfdefault}{bx}{n}

\usepackage{graphicx}
\usepackage{booktabs,tabularx}

\usepackage{xcolor}
\definecolor{flagcolor}{rgb}{0.80,0.55,0.00}

\usepackage{hyperref}
\usepackage{url}

\title{Causal Discovery Bench: Can LLMs Autonomously Conduct Effective Experiments?}

\author{Zhuoyang Liu \\
Independent Researcher \\
CA, USA \\
\texttt{liu.zhuoyang.22@gmail.com} \\
\And
Ning Peng \\
Independent Researcher \\
MA, USA \\
\texttt{npeng@umich.edu} \\
\AND
Ming Shen \\
Independent Researcher \\
CA, USA \\
\texttt{mshen1019@gmail.com} \\
\AND
Zijun Gao \\
University of South California \\
LA, USA \\
\texttt{zijungao@marshall.usc.edu}
}

\preprintfinalcopy
\begin{document}

\maketitle
\lhead{Preprint}

\begin{abstract}
As large language model (LLM) agents get actively adopted for autonomous research, they must not only reason over existing evidence, but also design informative experiments, allocate resources, analyze sequential observations, and decide when to stop and make conclusions. Yet these capabilities remain poorly measured. We introduce Causal-Discovery-Bench (\url{https://github.com/mshen1019/causal-discovery-bench}), a verifiable sandbox with configurable linear and nonlinear structural causal relationships and resource constraints to evaluate sequential experimental design, causal discovery, and treatment-effect estimation without domain-semantic cues. Across 6,727 episodes (3,802 linear and 2,925 nonlinear) with 13 models from 5 vendors, we analyze agent behavior along 40 metrics in seven families. Agents' performance exhibits a four-level difficulty hierarchy spanning single-experiment design, sequential exploration of causal relationships, resource-aware experimentation balancing cost and statistical power, and autonomous resolution of conflicting evidence. We identify a substantial performance gap between causal inference from correctly curated evidence and navigating causal discovery through autonomous experimentation: Analyzing agents' action trajectories attributes 94.4\% of attributed errors in the linear setting (84.5\% in the nonlinear setting) to experimental design rather than downstream inference, a finding further supported by our ablation studies. Moreover, among episodes with stored observational rows, agents had already collected sufficient data to identify the causal parents but failed to construct an estimator in 80\% of linear cases (68\% of nonlinear cases). While larger budgets primarily improve coverage without ensuring controlled experimentation in the linear setting, they increase controlled experiments and reduce dose–response estimation errors in the nonlinear setting. In contrast, basic statistical tools substantially improve performance across multiple model families, while advanced tools provide little additional benefit when models lack prerequisite lower-level capabilities. 

\end{abstract}

\section{Introduction}
\label{sec:introduction}

The fast growing capability of Large Language Model (LLM) agents has motivated strong interest in deploying them for autonomous scientific research, machine learning engineering agents, and recursive self-improvement (RSI). Such use cases require an LLM agent to actively navigate through research questions under limited budgets, turning observations into hypotheses,
hypotheses into experiments, and experimental results into both revised
conclusions and new directions of exploration. Generating evidence via autonomous sequential exploration, rather than merely reasoning over supplied evidence, becomes an increasingly sought-after capability.

\begin{center}
    \emph{Can an LLM agent self-navigate the design and analysis of a sequence of experiments using potentially limited resources to solve open questions?} 
\end{center}
Existing research reveals failures in interactive discovery and long-horizon autonomous research but does not diagnose agents' missing capabilities and failure modes when handling such tasks under finite data-acquisition budgets. We built \textsc{Causal-Discovery-Bench}, a verifiable sandbox to shed light on why LLM agents struggle and to design experiments that elicit how they can be strengthened without training. Running with configurable structural equation models and resource constraints, \textsc{Causal-Discovery-Bench} provides interactive environments for LLM agents to perform data-driven causal discovery tasks through sequential experiments and automatically evaluates their task performance against the ground truth. Variable names \textsc{Causal-Discovery-Bench} are deliberately randomized to rule out memory effects from domain-semantic cues, so that task performances reflect  agents' true causal discovery.

We find that the agents struggle to design sequential experiments to collect evidence rather than to interpret existing evidence, and
that this single deficit persists through every intervention that does not free agents from experiment design. Provided with adequate controlled experiment data, a mid-tier production model identifies the correct causal network structure in 14 of 15 sandboxes; however, when the same volume of data is collected but with inadequate variable control measuring total rather than direct treatment, the success number drops to 9. Across all 3,802 linear (2,925 nonlinear) episodes, 94.4\% (84.5\% for nonlinear) of attributed errors occur before the agent has constructed any comparison that identifies the effect. Increasing the resource budget by $125\times$ increases success rates in identifying causal network structure from 17\% to 78\% (nonlinear: 10\% to 73\%). However, in the linear setting, agents run no controlled experiments below the top budget and conduct at least one in only 15\% of episodes even there (nonlinear: 53\%), showing that improvements in linear accuracy are primarily driven by exploring more candidates rather than properly isolating their treatment effects.

In a separate study, a five-step procedure induces controlled experimentation in three models that never exhibit it unprompted, although it
does not close their final task performance gap with stronger models.

\section{Related Work}
\label{sec:related-work}
\paragraph{Autonomous research and long-horizon agents.}
LLM agents have been increasingly deployed to plan and execute scientific
experiments
\citep{boiko2023autonomous}. 
However, evaluations of autonomous research and
professional workflows reveal persistent difficulties in checking intermediate
results, pivoting from unsuccessful approaches, and maintaining task objectives
\citep{fei2026autoresearch,zhu2026workflowgym}. Agents can even underperform simple test-time scaling methods based on random retries
\citep{wang2026rethinking}. Interpreting these results is challenging because objective measures of research quality are often unavailable. Moreover, variation in agents’ domain knowledge can complicate standardized comparisons across models. We address these limitations with an interactive sandbox that provides known causal ground truth, removes domain-specific semantic cues, and supports controlled ablations of budgets, tools, and instructions.

\paragraph{Interactive model discovery and causal reasoning benchmarks.}
Several interactive benchmarks are relevant but differ in the evidence agents
receive, the resources they may spend, and the task itself: BoxingGym evaluates
predictive model discovery in domain-grounded environments
\citep{gandhi2025boxinggym}, CausalGame reasoning under hidden confounding,
selection bias and measurement error \citep{chen2026causalgame}, and IGDA edge
queries answered with binary feedback \citep{havrilla2025igda}. CausaLab is
closest, since it also uses hidden SCMs and budgeted interventions, but
evaluates broad graph and equation recovery alongside held-out prediction
\citep{yang2026causalab}. Our emphasis is diagnostic and prescriptive. We
isolate where an agent's experimental design, statistical power and causal
inference break down, and we test what repairs them. We build on CausalProfiler's SCM infrastructure for explicit assumptions, known ground truth, and sequential
interaction with a persistent population \citep{panayiotou2025causalprofiler}. Appendix~\ref{app:concurrent} sets out in detail what this study examines that the closest benchmarks, concurrent ones included, do not.

\section{Causal Discovery as Sequential Experimentation}
\label{sec:cd}
We evaluate LLM agents’ ability to design and analyze a sequence of experiments for causal discovery. Each task takes place in a sandbox that simulates an online A/B testing environment, where a population of instances produces a noisy outcome $Y$ influenced by a subset of variables $\mathbf{X}=(X_1,\ldots,X_p)$. Agents must use experimental evidence to identify which variables have a direct causal effect on $Y$ and estimate the magnitudes.

The causal structure is represented by a directed acyclic graph (DAG) $\mathbf{G}=(\mathbf{V},\mathbf{E})$, with vertices $\mathbf{V}=\{V_1,\ldots,V_{p+1}\}$ corresponding to $V_j=X_j$ for $j\leq p$ and $V_{p+1}=Y$. A vertex $V_i$ is an \textit{ancestor} of $V_j$ if a directed path leads from $V_i$ to $V_j$, and a \textit{parent} of $V_j$ if $(V_i,V_j)\in\mathbf{E}$. Each directed edge represents a direct causal effect. The discovery objective is therefore to identify the parents of $Y$.

Each sandbox contains a finite population of $N$ instances governed by a linear structural equation model (SEM) with instance-specific coefficients. Let $\mathrm{Pa}(j)$ denote the index set of the parents of $V_j$. For instance $n$, the structural equation is
\begin{equation}
V_j^n :=
\sum_{i\in\mathrm{Pa}(j)}
\big(a_{ij}+\delta_{ij}^n\big)V_i^n
+\varepsilon_j^n,
\tag{SCM}\label{eq:scm}
\end{equation}
where $a_{ij}\neq 0$ is the baseline coefficient shared across the population, $\delta_{ij}^n$ is an exogenous coefficient deviation, and $\varepsilon_j^n$ is additive exogenous noise. All coefficient deviations and additive noise terms are mutually independent across edges, vertices, and instances. For each edge $(i,j)$, the deviations $\{\delta_{ij}^n\}_{n=1}^N$ are independently drawn from an identical zero mean distribution; likewise, for each vertex $j$, the noise terms $\{\varepsilon_j^n\}_{n=1}^N$ are independently drawn from an identical zero mean distribution across instances. For a vertex with no parents, the equation reduces to $V_j^n=\varepsilon_j^n$.

At initialization, each instance independently receives a realization of the additive noise vector $\boldsymbol{\varepsilon}^n=(\varepsilon_1^n,\ldots,\varepsilon_{p+1}^n)$ and the coefficient deviations $\boldsymbol{\delta}^n:=\{\delta_{ij}^n:(V_i,V_j)\in\mathbf{E}\}$. Both remain fixed throughout the task and are reused whenever the instance is sampled for an observation or experiment. Together with the baseline coefficients and \eqref{eq:scm}, they determine the instance’s observational values and its outcomes under intervention.

Agents interact with the sandbox through a sequence of \textit{observations} and \textit{experiments}, subject to a fixed total budget $B$. Both actions incur costs linear in sample size: an observation of $m$ instances costs $c_o\cdot m$, while an experiment involving $m$ instances across both intervention groups costs $c_e\cdot m$, where $c_e>c_o>0$ are the respective per-instance costs. For each action, the agent selects a sample size $m$ and samples $m$ instances uniformly from the population. Let $n_1,\ldots,n_m$ denote their indices. An observation returns the sampled data $\big((X_1^{n_j},\ldots,X_p^{n_j},Y^{n_j})\big)_{j=1}^m$.
For an experiment, the agent additionally selects a variable index set $\mathcal{S}\subseteq\{1,\ldots,p\}$ and two intervention vectors $\mathbf{x}_1$ and $\mathbf{x}_2$, each of dimension $|\mathcal{S}|$. The sampled instances are randomly divided into two equally sized groups to respectively receive the interventions $\operatorname{do}((X_i)_{i\in\mathcal{S}}=\mathbf{x}_1)$ and $\operatorname{do}((X_i)_{i\in\mathcal{S}}=\mathbf{x}_2)$. Each intervention replaces the structural equations for the selected variables with their assigned values. The remaining variables and outcome are then computed using Equation~\ref{eq:scm} and each sampled instance’s fixed noise realization $\boldsymbol{\varepsilon}^n$ and $\boldsymbol{\delta}^n$. Computed results are shared to the agent. 

The agents' task is to identify the direct parents of $Y$ and estimate the direct causal effect of each parent. With $V_{p+1}=Y$, the target parent set is
\[
\mathcal{P}_Y
:=
\{j\in\{1,\ldots,p\}:(V_j,V_{p+1})\in\mathbf{E}\}.
\]For each $j\in\mathcal{P}_Y$, the target effect is the average change in $Y$ induced by a unit increase in $X_j$ while holding the remaining variables $\mathbf{X}_{-j}$ fixed:
\[
\theta_j(x;\mathbf{z})
:=
\mathbb{E}\!\left[Y\mid\operatorname{do}(X_j=x+1,\mathbf{X}_{-j}=\mathbf{z})\right]
-
\mathbb{E}\!\left[Y\mid\operatorname{do}(X_j=x,\mathbf{X}_{-j}=\mathbf{z})\right].
\]Here, the expectation is over the data-generating distribution of $\boldsymbol{\varepsilon}^n$ and $\boldsymbol{\delta}^n$ for $n=1,\cdots,N$. 
Under the linear SEM in \eqref{eq:scm} and the assumption $\mathbb{E}[\delta_{ij}^n]=0$, this controlled direct effect becomes independent of $x$ and $\mathbf{z}$ and equals the baseline coefficient, i.e.
\begin{equation}
    \theta_j(x;\mathbf{z})\equiv\tau_j=a_{j,p+1}.
\end{equation}
We extend the linear setting with six nonlinear function families governing direct effects on $Y$, modifying only the final DAG layer while keeping upstream relationships linear. Unlike constant linear effects, nonlinear effects vary with intervention values, requiring agents to estimate unknown dose--response curves rather than scalar coefficients. This introduces an additional effect-estimation challenge while retaining the original objective of identifying causal parents and estimating their effects through sequential experiments.

In our task, the agent must return an estimated parent set $\widehat{\mathcal{P}}_Y$ and an effect estimate $\widehat{\theta}_j$ for each selected parent, using a sequence of observations and experiments whose cumulative cost does not exceed the budget $B$. This is highly challenging due to the following considerations.

\subsection{Challenges in identifying direct parents and their treatment effects.}
To identify whether a variable $X_j$ is indeed a parent of $Y$, agents must design controlled experiments or select observational adjustment methods that isolate each candidate variable’s direct effect. Without prior knowledge of the parent set or the intervening causal structure \citep{pearl2009causality, cinelli2024controls}, this task is quite challenging, as the agent has to resolve several potential confusions.

\paragraph{Correlation vs Causality} The agent needs to distinguish between spurious correlation and genuine causality, which requires combining observations with experiments properly. For example, as long as a common ancestor is shared, $X_j$ and $Y$ can demonstrate a correlation without any causal relationship. In such cases, the agent needs to perform proper controlled experiments to rule out $X_j$ from $Y$'s ancestors.

\paragraph{Parents vs Ancestors} The agent also needs to distinguish the parents of $Y$ from the ancestors.
Changing $X_j$ from $x'$ to $x$ while leaving the other variables to follow their structural equations yields the total average treatment effect 
\begin{equation}
\tau_j(x,x')
:=
\mathbb{E}\big[Y\mid\operatorname{do}(X_j=x)\big]-
\mathbb{E}\big[Y\mid\operatorname{do}(X_j=x')\big]=
(x-x')\sum_{\pi:X_j\rightsquigarrow Y}
\prod_{(u,v)\in\pi}a_{uv}\footnote{\text{The $\{\delta_{ij}^n\}_{i,j,n}$ are zeroed by the expectations as they are mutually independent and each has zero mean.}}.
\label{eq:total}
\end{equation}
The total treatment effect can be misleading for identifying direct parents: it
is often nonzero as long as  $X_j$ is an ancestor $Y$. 

\paragraph{Direct vs Total Treatment Effects from Confounders} Even if the agent has correctly identified $X_j$ as a direct parent, it remains nontrivial to calculate the direct treatment effect if $X_j$ is a confounder shared with $Y$ and some other direct parent(s) of $Y$. In rare cases, such a confounder can have a zero total treatment effect on $Y$ if its direct treatment effect cancels out with its effect channeled through other direct parents. 

In addition to designing a sequence of observations and experiments that can, in principle, resolve these puzzles, agents must understand that they work in an environment where heterogeneous coefficients and noisy outcomes can introduce sampling uncertainty. They must quantify uncertainty in their effect estimates and choose sample sizes that provide sufficient statistical power for their inference within the available budget.

\subsection{Challenges from finite action budgets}
The finite budget makes causal discovery a sequential experimental design problem. Each action must be chosen using the accumulated evidence and the remaining resources. Agents face four interrelated decisions: which causal hypotheses to test next; how to design observation or experiment sets that identify direct effects; how many samples to allocate to each test given its cost and statistical power; and how to update causal conclusions and resolve conflicting evidence. These decisions require balancing the breadth of hypothesis testing against the precision of individual effect estimates, while determining whether further data collection is warranted before submitting a final conclusion.

\section{The Sandbox Design}

\begin{figure}[t]
\begin{center}
\includegraphics[width=\textwidth]{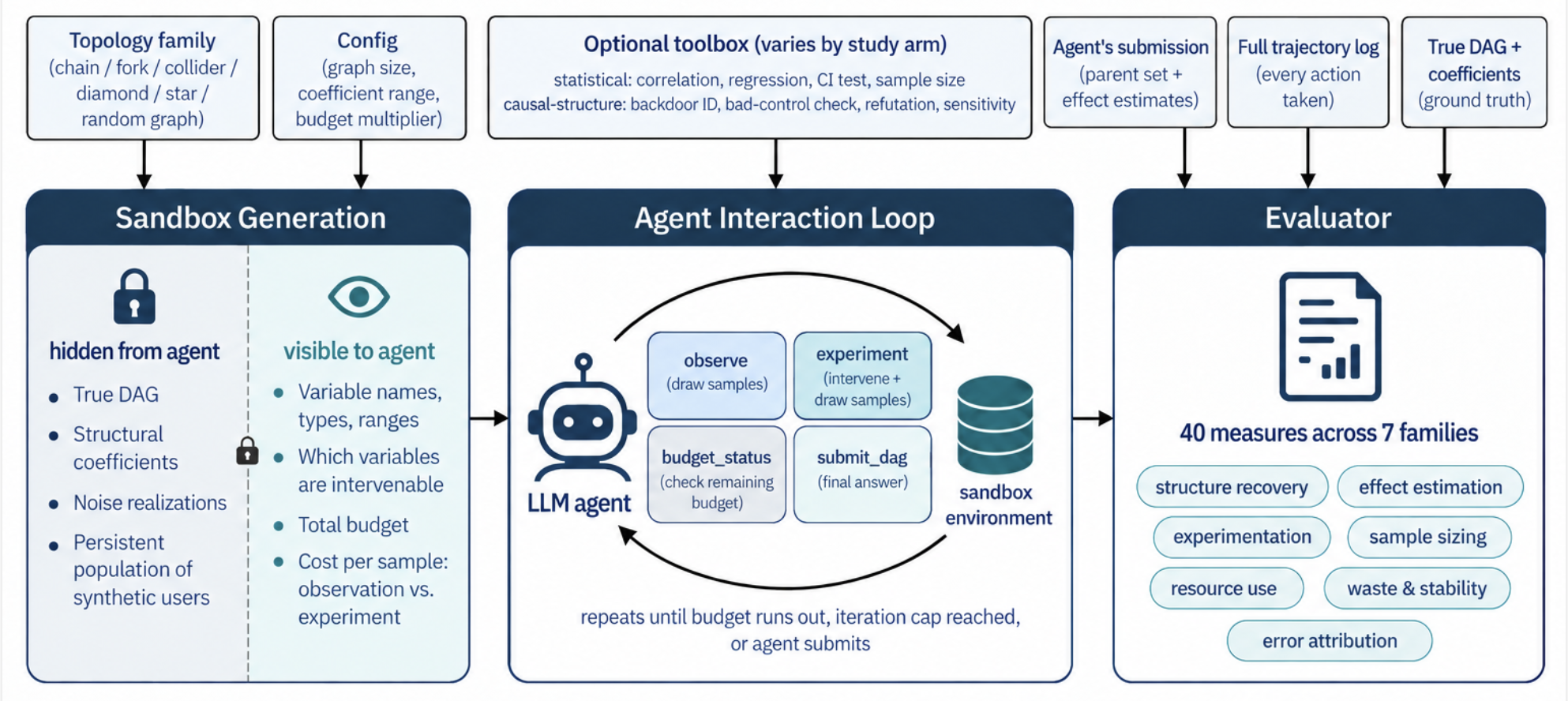}
\end{center}
\caption{The sandbox pipeline. Left: generation splits every instance into a hidden half (the true causal model) and a visible half (the interface the agent actually sees). Middle: the agent interacts through four priced core actions in a loop, optionally equipped with a statistical or
causal-structure toolbox depending on the study arm, never seeing the hidden half directly. Right: the evaluator scores the final submission and the full trajectory against the hidden ground truth, producing forty measures across seven families.}
\label{fig:sandbox-pipeline}
\end{figure}

The sandbox implements the sequential causal discovery task defined in Section~\ref{sec:cd}. Its design separates population generation, agent
interaction, and evaluation: the simulator generates observations and
experimental outcomes, the agent chooses how to collect and interpret evidence,
and the evaluator compares the resulting submission and trajectory against the
true causal model.

Figure~\ref{fig:sandbox-pipeline} shows the three stages end to end.

\subsection{Sandbox generation}
\label{sec:scm}
\begin{figure}[btp]
\begin{center}
\includegraphics[width=\textwidth]{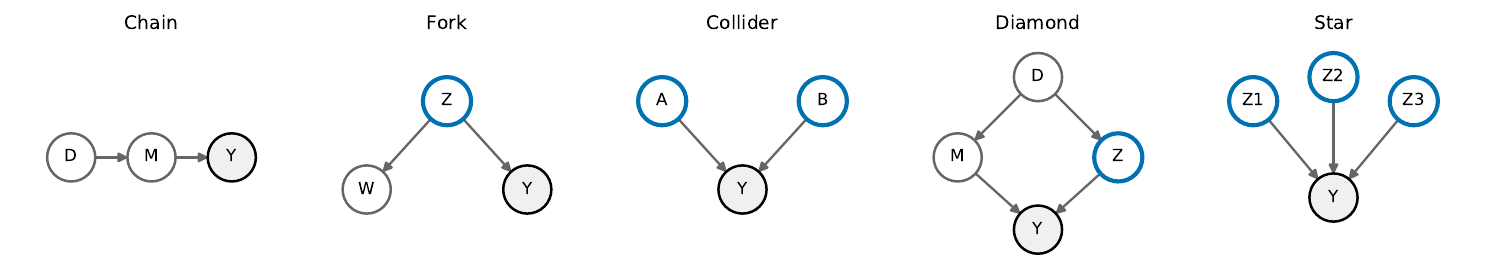}
\end{center}
\caption{Canonical examples for our five DAG families. Blue circles
are parents of the outcome $Y$}
\label{fig:topologies}
\end{figure}

We develop a diverse set of sandboxes to stress test the agents in distinctive ways to draw insights on their task performance traits and failure modes. Among these sandboxes, we consider five special families of DAGs with special topological structures: chain, star,
fork, collider, and diamond (Figure~\ref{fig:topologies}), to test whether the agent can handle particular reasoning and planning tasks. We also enrich our sandbox diversity with another family of DAGs generated by a random graph sampling procedure. We further consider different DAG sizes in each family to vary task difficulty. 

In our sandboxes, we don't model hidden causality links, i.e., all variables modeled in the SEMs can be observed by the agents. However, only a subset of the variables can be intervened with the \texttt{do} operator, with the subset being configurable for each sandbox. We also label variable names with randomly sampled strings that carry no domain semantic meaning so that our sandboxes measure the agents' raw causal reasoning powers without spillovers from memorized domain semantic cues that can be given away linguistically from variable names
\citep{zecevic2023parrots}.

Given the DAG structure, each sandbox extends CausalProfiler \citep{panayiotou2025causalprofiler} to initialize its parameters in closed-form SEMs and its persistent population. These parameters provide the evaluator with ground truth used to score agents' performances. We provide details on this initialization procedure in Appendix~\ref{app:sandbox}.

We additionally evaluate six nonlinear treatment-effect families, modifying only the final-layer relationships leading to $Y$ while retaining the original upstream linear structure. This isolates the added challenge of nonlinear effect estimation while preserving the original DAG structures and experimental framework. Unlike the linear setting, where the functional form is known by default, agents are informed only of the candidate nonlinear function families, not which one governs each relationship. They must therefore identify causal parents and estimate their unknown dose--response curves from experimental evidence.

We built another version to draw observation and experiment samples with replacement for benchmark robustness check. Study results between both samplers differ by less than 4\%. (Appendix~\ref{app:sampler}).

\subsection{Agent interface}
\label{sec:task}
The sandbox sends its task to an agent using a prompt in which the agent sees the data description that a machine-learning engineer would typically get when starting a project: The names and pointers to variables $\mathbf{X}$ and outcome $Y$, as well as their data type, range, and whether they can be intervened upon with the \texttt{do} operation.
The total budget for observation and experimentation actions and per sample costs are also provided. The DAG,
structural coefficients, and noise realizations remain hidden from the agent. The agent works on the task through multiple turns of interactions with the sandbox, where at any turn it can take any of the four actions: \texttt{observe} to see samples, \texttt{experiment} to intervene with samples and see raw results, \texttt{budget\_status} to check the remaining budget, and \texttt{submit\_dag} to conclude the task by submitting the parents of $Y$ and their per-unit treatment effects to the evaluator.
For nonlinear tasks, the agent additionally receives descriptions of the six candidate nonlinear function families but is not informed of the true functional form associated with each causal relationship. Consequently, the agent must infer both the relevant causal parents and the shapes of their dose--response relationships from observational and interventional evidence. The final submission is evaluated against the ground-truth nonlinear effect curves rather than only against scalar treatment-effect coefficients.
We log agents' full task trajectory, including their actions and outputs at each turn, and feed the trajectory to the evaluator.

\subsection{Evaluator and Metrics}
\label{sec:budget}
The evaluator assesses an agent's task performance using a comprehensive metric system that covers different dimensions of an
agent's ability to carry out a causal discovery task. The metric system examines the agent's entire task trajectories in a given set of sandboxes and evaluates not only the accuracy of final answers but also the progressive performances at each turn, including statistical power for sample sizes, budget efficiency, variable exploration coverage, etc. These dimensions are quantified by forty metrics that can be grouped into seven metric families
(further details in Appendix~\ref{app:metrics}). For nonlinear effects, we evaluate dose--response curves using normalized $L_1$ error over a fixed intervention grid, averaged across true causal parents, defined as
\begin{equation}
\mathcal{L}_{\mathrm{curve}} = \frac{1}{|\mathcal{P}_Y|}\sum_{j\in\mathcal{P}_Y}\frac{\frac{1}{K}\sum_{k=1}^{K}|\widehat{f}_j(x_k)-f_j(x_k)|}{\frac{1}{K}\sum_{k=1}^{K}|x_k|},
\label{eq:nonlinear_curve_error}
\end{equation}
where $f_j$ and $\widehat{f}_j$ denote the true and estimated effect curves, respectively. This metric reduces to the original scalar effect error for linear relationships.

The seventh family is what turns a wrong answer into a missing capability. It
charges every wrong edge back to the evidence the agent had gathered for that
candidate, and those states form four levels that the task requires in order:
(i) designing one experiment that identifies the effect, (ii) covering every
candidate, (iii) sizing each test to the power the budget can buy, and (iv)
resolving the evidence once it conflicts. The levels are ordered rather than merely distinct, because
failing at an earlier one hides the later ones: an agent that never tests a
candidate produces no evidence that could be misread, so its error can only be
charged to the level at which it stopped. An agent's profile over the four is
therefore read from the bottom up, and a level is only informative once the
ones below it are passed.

Our metric system also evaluates task difficulty based on budget scarcity by considering the ratio between available budgets and the heuristic budget $B_{\mathrm{test}}$ needed to complete the task using a strong algorithm, which consists of a screen followed by
one adequately powered controlled contrast per candidate without using
the true graph (details in Appendix~\ref{app:budget}). 
Particularly, we set a default budget allocation of \begin{equation}
    B=5B_{\mathrm{test}} \tag{Default Budget}\label{eq:default-budget}
\end{equation} for our sandbox and treat this budget as a baseline in later experiments.

\section{Experiments and Findings}
\label{sec:design}
\label{sec:experiments}


\subsection{Where can an agent fail}
\label{sec:exp-locus}
We run a set of experiments to understand why an agent can conclude with wrong parents or estimate treatment effects with poor accuracy. We focus on how \texttt{Claude Sonnet 4.6} and apply the budget rule at \ref{eq:default-budget} throughout the experiments. We summarize the analysis perspectives provided by these experiments in
Table~\ref{tab:failure-ablations}.
\begin{table}[h]
\centering
\small
\begin{tabularx}{\linewidth}{@{}lX@{}}
\toprule
\textbf{Ablation study} & \textbf{Detailed setup} \\
\midrule
\texttt{Estimation-only}
    & Estimating treatment effects given DAG; \texttt{15 sandboxes} \\
\texttt{Inference-only}
    & Identifying parents and estimating treatment effects given a set of observation and experiment data; \texttt{30 sandboxes}, \texttt{15 control + 15 treatment} \\
\texttt{DAG Topology}
    & Full causal discovery task on different DAG families; \texttt{318 sandboxes} \\
\texttt{DAG Sizes}
    & Full causal discovery task with different graph sizes (6-18 vertices) in each DAG family; \texttt{70 sandboxes} \\
\bottomrule
\end{tabularx}
\caption{Ablation studies used to localize agent failures.}
\label{tab:failure-ablations}
\end{table}

In the \texttt{estimation-only} experiment, we ablate the DAG-structure discovery component from our task by directly showing the DAG in each sandbox. Controlled interventions that separately vary each parent are sufficient for identification. However, \texttt{Claude Sonnet 4.6} did not conduct any controlled experiments in any of the 15 task trajectories we ran with it. This shows the agents' fundamental limitation in conducting controlled experiments.

In the \texttt{Inference-only} experiment, we ablate sequential experimentation entirely and provide a fixed evidence
package comprising one observation and one experiment dataset for each candidate parent variable, although the agent is still allowed to make additional observations and experiments. We evaluate two matched arms across the same 15 sandboxes. In the controlled arm, each candidate is subjected to a \texttt{do} operation with all other candidates held fixed, allowing the experiment to identify direct effects. In the
treatment arm, the focal candidate is still applied with a \texttt{do} operation, but the remaining variables
are uncontrolled, so the experiment measures total effects. The agent recovers the exact parent set in 14 of 15
controlled-evidence trajectories but only 9 of 15 treatment-evidence trajectories.

These two studies show that agents can usually reach the right conclusion when they are given the right
evidence. Once supplied with misleading (though true) evidence, the correct identification of parents drops from 14 to 9, signaling that the agent struggles to make independent adjustments regarding evidence quality. If required to design sequential experiments, the accuracy of treatment effect estimation drops for correctly identified parents, with the mean absolute error increasing from $0.11$ to $0.32$.
This result explains why 94.4\% of the errors in our broader evaluation over 3,802 agent task trajectory episodes occur before the final inference step. Appendix~\ref{app:metrics} reports this attribution separately for each model, and measures how much of it depends on counting only controlled experiments as identifying evidence. 

In the \texttt{DAG sizes} experiment, we keep the original causal discovery task fixed while increasing the size of random graphs from 6 to 18 variables. The agent fails to recover the exact parent set at every size, but
the source of difficulty changes systematically. The mean number of unresolved
candidates increases from $4.6$ to $8.5$, whereas recall among candidates the
agent resolves remains approximately constant ($0.48$ at six variables and
$0.44$ at eighteen, with no monotonic trend between them). Increasing the graph
size, therefore, primarily reduces how much of the problem the agent
investigates, rather than how accurately it judges the variables it encounters.

In the \texttt{DAG topology} experiment, we vary the graph across five canonical structures. Across 2,842 trajectories,
uncontrolled contrasts account for only $1\%$ of attributed errors in graphs
without mediation ($n=1{,}738$), but $44\%$ in graphs with mediation
($n=1{,}104$). Never-tested errors move in the opposite direction, decreasing
from $69\%$ to $42\%$. Graph topology, therefore, determines which capability
the task demands: mediation tests whether the agent uses appropriate controls,
whereas graphs without mediation primarily expose failures in hypothesis
coverage.

These two experiments separate two sources of difficulty. Graph size
determines how much of the task the agent leaves unfinished, while graph
topology determines which causal-reasoning skill is required to finish it.

\subsection{How models differ}
\label{sec:exp-models}
Running thirteen models in the same set of sandboxes with the same prompt, six of them on one fixed set of fifteen instances, we find both common and distinctive failure modes. Models demonstrate distinctive failure modes for more advanced experimental designs: one model makes errors mostly due to neglecting candidates from experimentation,
while another keeps running experiments without controlling variables. 

\begin{figure}[t]
\begin{center}
\includegraphics[width=\textwidth]{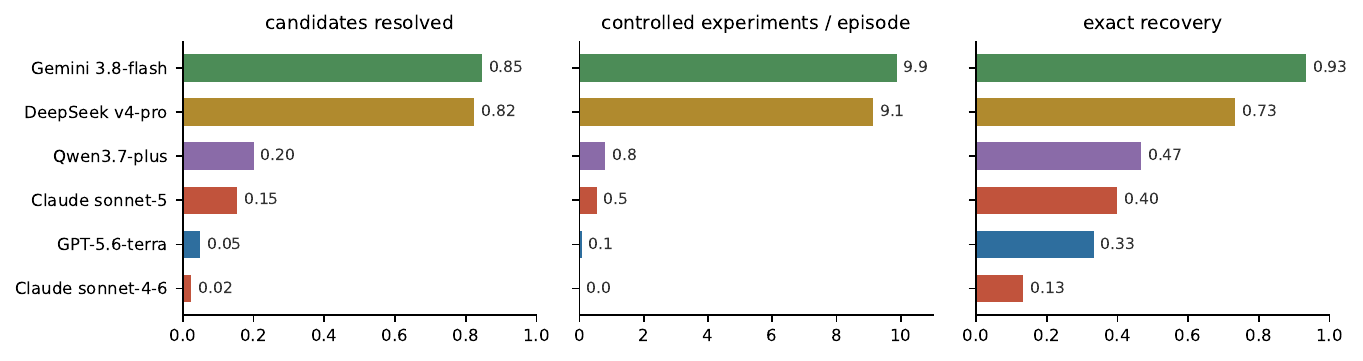}
\end{center}
\caption{Six models on one fixed set of fifteen instances, one bar each,
ordered by exact recovery. Left to right the panels follow the levels the task
requires in order: how many candidates the agent settles at all, how many
controlled experiments it runs, and how often it returns exactly the right
parent set. The two models that experiment run about ten controlled contrasts
per episode; the other four run at most one, with nothing in between. Colour
marks the vendor. The harder panel of Section~\ref{sec:exp-models} is left out
here: each of its models ran a different eleven-instance set, and three of
those instances carry outcome measures only, so its process rates are not
comparable.}
\label{fig:model-ladder}
\end{figure}

\paragraph{Neither model tier nor vendor strongly inform causal discovery capability.}
A flash model reaches 93\% exact recovery and finishes above every pro-tier
model in the panel, whereas another pro model reaches 73\%
(Figure~\ref{fig:model-ladder}, right). This ordering persists as we increase sandbox task difficulty, where the same flash model scores 91\% against 36\% for a
pro-preview model of its own family. Two vendors appear at both ends of the
ordering, so neither the parameter tier nor the model family predicts task performance.

\paragraph{Controlled experimentation distinguishes task performance.}
Two models conduct about ten controlled experiments per episode, whereas the other four conduct at most one; no model falls between these extremes (Figure~\ref{fig:model-ladder}, center). Among the 40 measures in Appendix~\ref{app:metrics}, only this behavior effectively divides the models into strong vs weak groups. Accuracy, recall, and budget use, instead, vary continuously. Final accuracy also obscures how answers were reached: the two weakest models resolve only 2\% and 5\% of candidate relationships, yet achieve exact parent-set accuracies of 13\% and 33\%, respectively. They can therefore reach correct answers without collecting the identifying evidence needed to support them.

\subsection{How to Fix Agents}
\label{sec:exp-levers}
We tried four ways of improving the agent without changing the model: more
budget, analysis tools, a tighter interface, and a written procedure. The two that supply resources leave its behavior where it was; the two that
supply the planning step change it, and still leave it short of a model that
plans for itself.
\begin{figure}[h]
\begin{center}
\includegraphics[width=\textwidth]{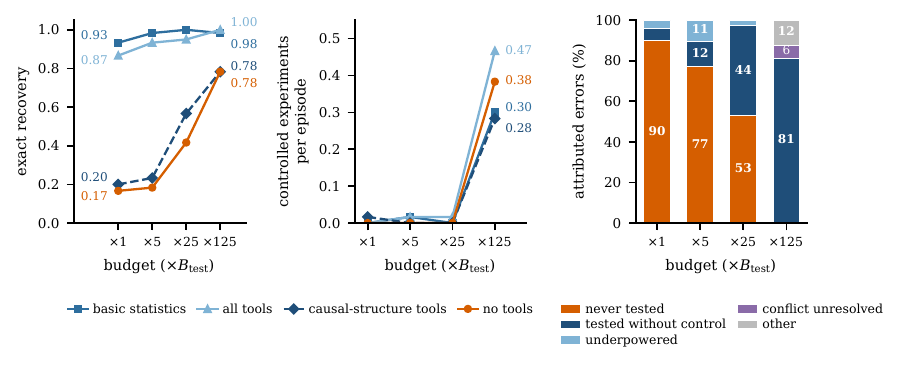}
\end{center}
\caption{\textbf{Left:} exact recovery against budget for each toolbox, on the
same instances. Four statistical primitives at $\times 1$ already exceed no
tools at $\times 125$, while the causal-structure toolbox tracks no tools at
both ends of the range. \textbf{Centre:} controlled experiments per episode
under the same conditions; no arm reaches half of one at any budget.
\textbf{Right:} where the remaining errors sit for the no-tools arm. Budget
does not remove the failure, it moves it from never-tested to
tested-without-control. The top budget leaves only 16 attributed errors, so the
shares in that column rest on a small count.}
\label{fig:budget-tools}
\end{figure}

\paragraph{Budget and tools buy more testing, not better testing.}
A budget 125 times larger lifts exact recovery from 17\% to 78\%, while the
agent runs no controlled experiment at all below the top budget, and one in only
15\% of episodes even there, against about nine per episode for a model that runs them by habit (Figure~\ref{fig:budget-tools}). What
changes is which mistake the agent makes: errors from never testing a candidate
fall from 90\% of the total to 0\%, and errors from testing one without holding
the others fixed rise from 6\% to 81\%. A statistical toolbox reaches the same
accuracy for a fraction of that budget by reading cheap observational rows
rather than buying interventions, and leaves the same contrast undone. A
causal-structure toolbox changes nothing at either end of the range: it checks
conclusions, and the agent has not produced one to check.

\begin{figure}[t]
\begin{center}
\includegraphics[width=0.58\textwidth]{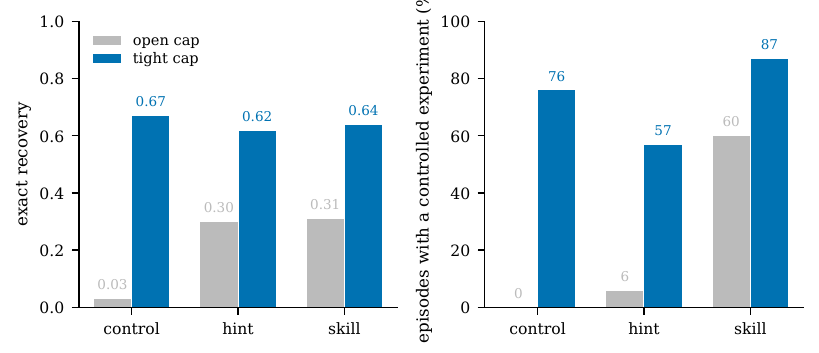}
\end{center}
\caption{Cap against prompt, on the instances of the preregistered $3\times2$.
Under the open cap the three prompts are far apart; under a cap set to the
instance's power requirement they are indistinguishable and every one of them
runs controlled experiments. The environment, not the wording, decides whether
the behavior appears.}
\label{fig:cap-prompt}
\end{figure}

\paragraph{Capping the purchase size makes the agent test properly.}
Capping how much a single call may buy raises exact recovery from 21\% to 64\%
and takes the share of episodes containing a controlled experiment from 22\% to
73\%; under the tight cap the three prompts become indistinguishable
(Figure~\ref{fig:cap-prompt}). The cap works by choosing the sample size the
agent will not choose for itself, since recovery peaks at a cap of 7 against a
power requirement of 6.58. A written five-step procedure has the same
effect without touching the interface, taking three models from no controlled
experiments at all to between 2.7 and 4.5 per episode. Their accuracy still lands at 87\%, 73\% and 40\%, short of the 100\% reached by
GPT-6 Astra, the strongest of the three models that run controlled experiments
without being told to; the other two reach 60\% and 53\%. The behavior is installable; the capability is not.

Budget buys coverage: more candidates tested, none of them settled. The only
levers that buy the contrast itself are the two that take a decision away from
the agent, the cap by fixing the sample size and the procedure by stating the
comparison, while the toolbox that checks conclusions never meets one.

\subsection{Do the findings generalize to nonlinear effects?}

Across 2,925 nonlinear episodes with 13 models and six treatment-effect families, our findings largely corroborate the linear results. Experimental design accounts for 84.5\% of attributed errors (94.4\% linear), while 68\% of eligible episodes lack an estimator despite sufficient observational evidence (80\% linear). Detailed experimental configurations and results are provided in Appendix~\ref{app:nonlinear-results}.

Increasing budgets from $1\times$ to $125\times$ reduces median normalized curve $L_1$ error (Eq.~\ref{eq:nonlinear_curve_error}) from 0.1866 to 0.0534 and increases median controlled experiments from 4 to 14, indicating stronger budget effects on controlled experimentation than in the linear setting.

Strategy guidance increases the controlled-experiment rate from 0.54 to 0.97 and reduces median curve error from 0.1017 to 0.0779, yet guided weaker models still underperform stronger unguided models. A targeted shape-fitting tool also outperforms both no optional tools (0.0761 vs. 0.0941) and all sixteen tools (0.1017). Thus, effective experimental design and tool use remain central across both settings.

\section{Conclusion}
We develop \textsc{Causal-Discovery-Bench}, a sandbox in which an agent is required to
find the direct causes of an outcome by buying observations and interventions
against a finite budget. Every action it takes is recorded and scored on forty measures, and every cause
it gets wrong is traced to the step it failed at: designing an experiment that
identifies the effect, covering every candidate, sizing each test to the power
it needs, or resolving evidence that conflicts. Across 6,727 linear and nonlinear episodes from thirteen models, experimental design accounts for most attributed errors (94.4\% linear; 84.5\% nonlinear). Agents resolve the evidence they hold and fail to produce evidence worth holding. Despite sufficient observational evidence, agents fail to construct estimators in 80\% of eligible linear episodes (68\% nonlinear).

Our analysis identifies the following failure patterns and potential fixes.
Causal inference needs a controlled comparison, varying one candidate while
holding the rest fixed. More budget and better tools get the agent to test more
candidates, and it still never holds the others fixed while it tests one. Two things do make it hold them fixed: a cap on how much one purchase may buy,
and a written procedure in the prompt. The cap supplies the sample size the agent
does not compute, and the procedure states the comparison it does not
construct. Neither, however, makes the agent as accurate as a
model that needs no such help, and that gap in accuracy is visible only in what
the agent did to reach its answer. Our nonlinear experiments largely confirm these findings, while showing that larger budgets can also promote controlled experimentation.

\paragraph{Limitations.}
Every variable in the sandbox is observed. If the agent can intervene on a
candidate, a hidden cause does no harm, because the intervention cuts the path
through it and the contrast still recovers the direct effect. If it cannot,
adjustment is its only option, and a hidden cause biases the adjustment. The
agent would then have no way to get that candidate right, so an error there
would not be its fault. Many discovery problems in practice do contain hidden
causes, and reaching them is work we have not done.

Rather than only evaluating the final answer an agent submits, our method reads a
signal off the trajectory by charging for every action and recording which of
the four steps an error came from. That record tells an agent whose evidence
supports the causes it names from one whose evidence does not. Nothing
in it depends on causal graphs, so the same scoring applies past direct-cause
discovery to any task where evidence has to be bought before it can be read. As
agents are handed more research that requires collecting new evidence, that is
the sandbox worth building, one that prices every action and reads the
trajectory against the four capabilities the task demands in order.

\subsection*{AI use statement}

(This section is \textbf{required} and does not count toward the page limit.)

We used generative AI tools (Claude, Anthropic; Codex, OpenAI) for tasks that require
disclosure: implementing and refining our experiment infrastructure of Appendix~\ref{app:pipeline}; processing unstructured experiment data to draw insights, such as in Section~\ref{sec:exp-levers}; writing the offline analysis
and plotting scripts that recompute reported metrics from stored trajectories and render Figures~\ref{fig:topologies}--\ref{fig:cap-prompt}; producing the pipeline schematic of Figure~\ref{fig:sandbox-pipeline}, whose content and layout the authors specified and checked against the implementation; and helping to interpret results, including several
explanatory paragraphs in Section~\ref{sec:exp-levers}. We did not use them to
generate synthetic data, to build theoretical or conceptual frameworks, to
formulate mathematical claims, or to write proofs: the data-generating process,
the reference cost in Appendix~\ref{app:budget}, and the derivations in
Appendix~\ref{app:proofs} are the authors' own. Translation and thematic
analysis do not apply to this work.

Under recommended disclosure, we also used them to draft and edit parts of
Sections~\ref{sec:scm} and~\ref{sec:experiments}, convert text to \LaTeX, and
search for related work. We verified AI-suggested citations against the sources
themselves. We
take responsibility for the final content of this work.

\bibliography{references}

@misc{fei2026autoresearch,
  title = {How Do Agents Fail on {AutoResearch}: End-to-End Diagnostic Evaluation on 100 Real-World Frontier Research Tasks},
  author = {Fei, Yanlin and Liu, Nazhou and Yu, Xinmiao and Chen, Shaolong and Li, Lei and Thapa, Rahul and Ciobanu, Madalina and Singh, Navan Preet and Mao, Qingqing and Das, Ritankar},
  year = {2026},
  eprint = {2608.14905},
  archivePrefix = {arXiv},
  primaryClass = {cs.CL},
  doi = {10.48550/arXiv.2608.14905},
  url = {https://arxiv.org/abs/2608.14905}
}

@misc{chen2026causalgame,
  title = {{CausalGame}: Benchmarking Causal Thinking of {LLM} Agents in Games},
  author = {Chen, Zhenhao and Chen, Yongqiang and Liu, Chenxi and Yu, Junchi and Song, Xiangchen and Li, Zijian and Li, Jialin and Torr, Philip and Han, Bo and Zhang, Kun},
  year = {2026},
  eprint = {2607.04293},
  archivePrefix = {arXiv},
  primaryClass = {cs.AI},
  doi = {10.48550/arXiv.2607.04293},
  url = {https://arxiv.org/abs/2607.04293}
}

@misc{zhu2026workflowgym,
  title = {{Workflow-GYM}: Towards Long-Horizon Evaluation of Computer-use Agentic tasks in Real-World Professional Fields},
  author = {Zhu, Liya and Ding, Jingzhe and Zhang, Jian and Xue, Jianbo and Liang, Shihao and Zhang, Ge and Zhu, Yi and Zeng, Duju and Gao, Xiang and Gu, Qingshui and Gao, Mailun and Che, Huimin and Zhao, Yan and Zhou, Peiheng and Wang, Haojun and Xian, Chaobo and Le, Lili and Wu, Chi and Liu, Yiwei and Long, Shengda and Yang, Jiale and Xu, Fangzhi and Wu, Sijin and Duan, Haodong and He, Chao and Li, Zhaojian and Wang, Minchao and Zhou, Huan and Hou, Jiani and Yu, Chuqian and Shi, Weiran and Gao, Hongwan and Chen, Jiamin and Chen, Guanhong and Luo, Tingqin and Zhang, Kaiyuan and Yao, Zhixin and Hua, Qing and Jiang, Yuhao and Chen, Jin and Chen, Pu and Hu, Zhenyu and Li, Xingyu and Jiang, Zhengxuan and Cao, Meng and Long, Tianfeng and Wang, Haozhe and Wang, Mingzhang and Zhang, Yichen and Dai, Yiming and Zhang, Chenchen and Wang, Jiaying and Liu, Xinying and Liu, Xingzu and Zhang, Lingling and Chen, Xinjie and Qin, Yujia and Zhou, Wangchunshu and Wu, Zhiyong and Liu, Yang and Liu, Jiaheng and Zhang, Lei and Yan, Shen and Huang, Wenhao and Wang, Zaiyuan and Chang, Xiaolong},
  year = {2026},
  eprint = {2606.11042},
  archivePrefix = {arXiv},
  primaryClass = {cs.AI},
  doi = {10.48550/arXiv.2606.11042},
  url = {https://arxiv.org/abs/2606.11042}
}

@misc{gandhi2025boxinggym,
  title = {{BoxingGym}: Benchmarking Progress in Automated Experimental Design and Model Discovery},
  author = {Gandhi, Kanishk and Li, Michael Y. and Goodyear, Lyle and Li, Louise and Bhaskar, Aditi and Zaman, Mohammed and Goodman, Noah D.},
  year = {2025},
  eprint = {2501.01540},
  archivePrefix = {arXiv},
  doi = {10.48550/arXiv.2501.01540},
  url = {https://arxiv.org/abs/2501.01540}
}

@misc{yang2026causalab,
  title = {{CausaLab}: A Scalable Environment for Interactive Causal Discovery Toward {AI} Scientists},
  author = {Yang, Junlin and Zhang, Dylan and Song, Xiangchen and Dai, Qirun and Liu, Xiao and Chen, Yuen and Vashishtha, Aniket and Shi, Jing and Tan, Chenhao and Peng, Hao},
  year = {2026},
  eprint = {2605.26029},
  archivePrefix = {arXiv},
  primaryClass = {cs.AI},
  doi = {10.48550/arXiv.2605.26029},
  url = {https://arxiv.org/abs/2605.26029}
}

@misc{panayiotou2025causalprofiler,
  title = {{CausalProfiler}: Generating Synthetic Benchmarks for Rigorous and Transparent Evaluation of Causal Machine Learning},
  author = {Panayiotou, Panayiotis and Poinsot, Audrey and Leite, Alessandro and Chesneau, Nicolas and Schoenauer, Marc and {\c{S}}im{\c{s}}ek, {\"O}zg{\"u}r},
  year = {2025},
  eprint = {2511.22842},
  archivePrefix = {arXiv},
  doi = {10.48550/arXiv.2511.22842},
  url = {https://arxiv.org/abs/2511.22842}
}

@inproceedings{wang2026rethinking,
  title={Rethinking the evaluation of harness evolution for agents},
  author={Wang, Yike and Zhu, Huaisheng and Hu, Zhengyu and Yuan, Yige and Chen, Zhengyu and Senthil, Shakti and Hajishirzi, Hannaneh and Tsvetkov, Yulia and Dasigi, Pradeep and Xiao, Teng},
  booktitle={COLM 2026 The 2nd Workshop on Lifelong Agents: Learning, Aligning, and Evolving},
  year={2026}
}

@book{embretson2000irt,
  title     = {Item Response Theory for Psychologists},
  author    = {Embretson, Susan E. and Reise, Steven P.},
  year      = {2000},
  publisher = {Lawrence Erlbaum Associates}
}

@inproceedings{ghassami2018budgeted,
  title     = {Budgeted Experiment Design for Causal Structure Learning},
  author    = {Ghassami, AmirEmad and Salehkaleybar, Saber and Kiyavash, Negar and Bareinboim, Elias},
  booktitle = {Proceedings of the 35th International Conference on Machine Learning},
  series    = {PMLR},
  volume    = {80},
  year      = {2018}
}

@inproceedings{kocaoglu2017cost,
  title     = {Cost-Optimal Learning of Causal Graphs},
  author    = {Kocaoglu, Murat and Dimakis, Alexandros G. and Vishwanath, Sriram},
  booktitle = {Proceedings of the 34th International Conference on Machine Learning},
  series    = {PMLR},
  volume    = {70},
  year      = {2017}
}

@book{cohen1988power,
  title     = {Statistical Power Analysis for the Behavioral Sciences},
  author    = {Cohen, Jacob},
  edition   = {2nd},
  publisher = {Lawrence Erlbaum Associates},
  year      = {1988}
}

@article{colombo2014order,
  title   = {Order-Independent Constraint-Based Causal Structure Learning},
  author  = {Colombo, Diego and Maathuis, Marloes H.},
  journal = {Journal of Machine Learning Research},
  volume  = {15},
  pages   = {3741--3782},
  year    = {2014}
}

@article{welch1947generalization,
  title   = {The Generalization of {S}tudent's Problem when Several Different Population Variances are Involved},
  author  = {Welch, B. L.},
  journal = {Biometrika},
  volume  = {34},
  number  = {1--2},
  pages   = {28--35},
  year    = {1947}
}

@book{pearl2009causality,
  title     = {Causality: Models, Reasoning, and Inference},
  author    = {Pearl, Judea},
  edition   = {2nd},
  publisher = {Cambridge University Press},
  year      = {2009}
}

@article{cinelli2024controls,
  title   = {A Crash Course in Good and Bad Controls},
  author  = {Cinelli, Carlos and Forney, Andrew and Pearl, Judea},
  journal = {Sociological Methods \& Research},
  volume  = {53},
  number  = {3},
  pages   = {1071--1104},
  year    = {2024}
}

@article{zecevic2023parrots,
  title   = {Causal Parrots: Large Language Models May Talk Causality But Are Not Causal},
  author  = {Ze{\v{c}}evi{\'c}, Matej and Willig, Moritz and Dhami, Devendra Singh and Kersting, Kristian},
  journal = {Transactions on Machine Learning Research},
  year    = {2023}
}

@article{efron1979bootstrap,
  title   = {Bootstrap Methods: Another Look at the Jackknife},
  author  = {Efron, Bradley},
  journal = {The Annals of Statistics},
  volume  = {7},
  number  = {1},
  pages   = {1--26},
  year    = {1979}
}

@article{boiko2023autonomous,
  title   = {Autonomous Chemical Research with Large Language Models},
  author  = {Boiko, Daniil A. and MacKnight, Robert and Kline, Ben and Gomes, Gabe},
  journal = {Nature},
  volume  = {624},
  pages   = {570--578},
  year    = {2023}
}

@misc{havrilla2025igda,
  title         = {{IGDA}: Interactive Graph Discovery through Large Language Model Agents},
  author        = {Havrilla, Alex and Alvarez-Melis, David and Fusi, Nicolo},
  year          = {2025},
  eprint        = {2502.17189},
  archivePrefix = {arXiv}
}

@article{shimizu2006lingam,
  title   = {A Linear Non-{G}aussian Acyclic Model for Causal Discovery},
  author  = {Shimizu, Shohei and Hoyer, Patrik O. and Hyv{\"a}rinen, Aapo and Kerminen, Antti},
  journal = {Journal of Machine Learning Research},
  volume  = {7},
  pages   = {2003--2030},
  year    = {2006}
}
\bibliographystyle{preprint_style}

\appendix

\section{Deferred derivations}
\label{app:proofs}

\paragraph{Every ancestor has a non-zero total effect.}
Two directed paths from $X_j$ to $Y$ use distinct edge sets, so no monomial in
the sum of \eqref{eq:total} cancels another and the sum is a non-zero polynomial
in $\{a_{uv}\}$ whenever at least one directed path exists. The zero set of a
non-zero polynomial has Lebesgue measure zero, and the generator draws each
$a_{uv}$ from a continuous range, so the event that \eqref{eq:total} vanishes
for an ancestor of $Y$ has probability zero. This is the cancellation that
faithfulness rules out by assumption; here the sampling scheme rules it out
instead.

\paragraph{The two routes, and the cost of one conclusive contrast.}
Pinning $X_{-j} = \mathbf{X}\setminus\{X_j\}$ to zero removes every path but the
edge itself, so that
\begin{equation}
\Delta_j \;:=\; \tfrac{1}{2}\Big[
\mathbb{E}\big(Y \mid do(X_j{=}{+}1,\, X_{-j}{=}0)\big) -
\mathbb{E}\big(Y \mid do(X_j{=}{-}1,\, X_{-j}{=}0)\big)\Big]
\;=\; a_{jY} ,
\label{eq:delta}
\end{equation}
and $\Delta_j = 0$ iff $X_j \notin \mathrm{Pa}(Y)$; only a variable that is both
a descendant of $X_j$ and an ancestor of $Y$ has to be neutralised, since
anything else cancels across the arms. Because the generator places no variable
outside $\mathbf{X}\cup\{Y\}$, the same quantity is also identified by
conditioning: a set of observed variables can block every back-door path, and
regressing $Y$ on $X_j$ given that set returns $a_{jY}$ from observational rows
at $c_{\mathrm{obs}}$ each. Where a variable that must be neutralised cannot be
manipulated, only the second route is open, which is why we also report recall
restricted to intervenable parents.

Pinning leaves the residual variance of $Y$ in \eqref{eq:delta} exactly
$\sigma^{2}=\mathrm{Var}(\varepsilon_Y)$. Sizing a two-arm contrast to detect a
difference of $\delta_{\min}$ at $80\%$ power and $\alpha = 0.05$ gives
$n^{\ast}=16\sigma^{2}/\delta_{\min}^{2}$ rows per arm \citep{cohen1988power},
where $\delta_{\min}$ is the smallest coefficient the generator can draw. The
two arms sit at $\pm 1$, so the difference actually to be detected is
$2\delta_{\min}$ and this allocation is deliberately conservative. A conclusive
test therefore costs $2n^{\ast}c_{\mathrm{exp}}$ against total budget.

\paragraph{Why recovering the whole graph from $P(V)$ is not available.}
A linear model with non-Gaussian noise is identifiable from $P(V)$ alone by
independent-component methods \citep{shimizu2006lingam}, and the exogenous noise
here is uniform. That route is nonetheless closed: each row carries its own
per-edge coefficient deviation (Section~\ref{sec:scm}), so the population is not
a single fixed linear model and the identifiability result does not apply to it.
Even where it did, recovering $\mathbf{G}$ would leave every $a_{jY}$
unestimated, and the rows it reads are charged against total budget like
any others.

\section{The budget formula}
\label{app:budget}

\paragraph{The formula.}
Collecting the two phases gives the reference cost the body refers to,
\begin{equation}
B_{\mathrm{test}}
:=
10(p+1)c_o
+
\min(2k,p)\,
\frac{16\sigma_{\mathrm{ref}}^2}{\delta_{\min}^2}\,c_e ,
\label{eq:reference-budget}
\end{equation}
whose first term buys the observational screen over $p$ candidates and the
outcome, and whose second buys one adequately powered controlled contrast for
each of $\min(2k,p)$ of them. Every instance is priced by
\eqref{eq:reference-budget} and given $B=5B_{\mathrm{test}}$ by default.

\paragraph{Derivation.}
Prior work on budgeted experiment design fixes a budget and asks what structure
is learnable inside it \citep{kocaoglu2017cost, ghassami2018budgeted}; we invert
the question and price the strategy a competent agent who does not already know
the graph would have to run. It proceeds in two phases. It first screens: on $p$
variables it buys $10p$ observational rows and runs a constraint-based search
over them \citep{colombo2014order}. It then tests each surviving candidate with
one two-arm intervention powered at $80\%$, which by Cohen's formula needs
$n^{\ast} = 16\sigma^{2}/\delta_{\min}^{2}$ rows \citep{cohen1988power}, where
$\delta_{\min}$ is the smallest coefficient the generator can draw and
$\sigma^{2}$ is the noise on the edge under test, the variance left once the
other candidates are held fixed, taken at its largest over the instance.
Because a constraint-based screen on finite samples returns false neighbors, we
let it hand back up to $\min(2k, p-1)$ candidates.

\paragraph{What the formula does and does not price.}
It prices the \emph{controlled} route, because $\sigma^{2}$ is measured with the
other candidates held fixed; an agent that intervenes without controlling also
carries the variance the other parents contribute, so the same power costs it
more. It prices one \emph{test} rather than one instance, because $80\%$ power
per candidate is not $80\%$ power per instance: with several candidates to
settle, a multiplicity correction stands between the two. Across the benchmark
the median $B_{\mathrm{test}}$ is $31.7$ (IQR $23.8$--$48.5$) and one powered
contrast takes a median of $6.6$ rows.

\paragraph{The reference statistician.}
The procedure is: screen, one controlled contrast per candidate, Welch $t$-test
\citep{welch1947generalization}, Bonferroni over the candidate tests. Unlike the
pricing formula it uses no knowledge of $k$ and tests every candidate, so it may
run more tests than $B_{\mathrm{test}}$ pays for. Run on 15 instances with 20
sampling repetitions each, it recovers the exact parent set in 296 of 300
trials at the default $5B_{\mathrm{test}}$ and in 79 of 300 at
$B_{\mathrm{test}}$, which is what calibrates the default: a competent
strategy succeeds there and succeeds only about a quarter of the time
at the lower figure. Because the procedure is
conservative in the sense just described, both figures understate what the
budget affords.

\paragraph{Scripted strategies and the agent.}
Conditioning on observational rows solves the task as well as the procedure
above and at a fifth of its budget, so the comparison below places both beside
the agent on the same five canonical instances.

\begin{center}
\small
\begin{tabular}{llr}
\toprule
\textbf{Strategy} & \textbf{Budget} & \textbf{Exact recovery} \\
\midrule
conditioning on observational rows & $B_{\mathrm{test}}$            & $98.7\%$ (296/300) \\
conditioning on observational rows & $5B_{\mathrm{test}}$ (default) & $100\%$ (300/300) \\
LLM agents, six models pooled      & $5$ to $43.6\,B_{\mathrm{test}}$, median $8.6$ & $33.3\%$ (16/48) \\
\bottomrule
\end{tabular}
\end{center}

The agent row is not budget-matched, and the mismatch favours the agent.
The agent episodes ran under configurations that fix the total budget while
difficulty varies, which is why their budgets differ from one another and from
the rows above. With more budget than the scripted procedure needs, the agents
still recover the parent set in a third of episodes. The row covers 48 episodes
over five topologies, so we offer it as an illustration rather than a powered
comparison.

\paragraph{The variance proxy.}
$\sigma^2_{\mathrm{ref}}$ in \eqref{eq:reference-budget} is the largest
empirical variance of the per-unit coefficient deviations on an edge into $Y$,
plus the variance of the optional additional outcome-noise term. It normalizes
the budget across instances; it is not the sampling variance of any particular
experimental design, and \eqref{eq:reference-budget} is correspondingly a
reference price rather than a sufficient allowance.

\paragraph{Which variance enters where.}
Three different quantities go by the name variance in this paper, and the
table below separates them. None of the three involves a hidden or a fitted
parameter.

\begin{center}
\small
\begin{tabularx}{\linewidth}{@{}lX@{}}
\toprule
\textbf{Used for} & \textbf{Quantity} \\
\midrule
the power formula $n^{\ast}=16\sigma^{2}/\delta_{\min}^{2}$
  & $\sigma^{2}=\mathrm{Var}(\varepsilon_Y)$, the variance left in $Y$ once
    every other parent has been pinned, as derived in
    Appendix~\ref{app:proofs} \\
normalizing the budget across instances
  & $\sigma^{2}_{\mathrm{ref}}$, defined in the paragraph above \\
diagnostic labels and the per-call purchase cap
  & the empirical standard error of the agent's own inverse-variance pooled
    estimate, compared against the fixed constants $|t|>2.5$,
    $\delta_{\min}=0.5$ and an effect gate of $0.25$ \\
\bottomrule
\end{tabularx}
\end{center}

\section{Models}
\label{app:models}
Every model is the vendor's hosted endpoint, called with default sampling
settings and no system prompt beyond the task prompt of
Section~\ref{sec:task}. Names are as the vendor published them at the time of
the run; endpoints for several have since been retired, so the model string in
the released episode records is authoritative.

\begin{center}
\small
\begin{tabular}{lll}
\hline
\textbf{Model} & \textbf{Model string} & \textbf{Studies} \\
\hline
Claude Sonnet 4.6 & \texttt{claude-sonnet-4-6} & single-model studies, panels \\
Claude Sonnet 4.6, thinking & \texttt{claude-sonnet-4-6-thinking} & thinking ablation \\
Claude Sonnet 5 & \texttt{claude-sonnet-5} & workhorse panel \\
Claude Opus 5 & \texttt{claude-opus-5} & frontier panel \\
Claude Fable 5.1 & \texttt{claude-fable-5-1} & frontier panel \\
GPT-5.6 Terra & \texttt{gpt-5.6-terra} & scaffolding cross, panels \\
GPT-5.6 Sol & \texttt{gpt-5.6-sol} & frontier panel \\
GPT-6 Astra & \texttt{gpt-6-astra} & scaffolding cross, panels \\
Gemini 3.8 Flash & \texttt{gemini-3.8-flash} & scaffolding cross, panels \\
Gemini 3.1 Pro Preview & \texttt{gemini-3.1-pro-preview} & frontier panel \\
DeepSeek V4 Pro & \texttt{deepseek-v4-pro} & scaffolding cross, panels \\
Qwen3.7 Plus & \texttt{qwen3.7-plus} & scaffolding cross, panels \\
Qwen3.8 Max & \texttt{qwen3.8-max} & frontier panel \\
\hline
\end{tabular}
\end{center}

\section{Instance construction}
\label{app:sandbox}

\paragraph{The answer key and the realized population.}
The evaluator reads the parent set from the DAG and each direct effect from the
baseline SEM coefficient, so the key is exact by construction. The average
effect realized in a finite population can still differ from that target,
because the per-unit coefficient deviations of \eqref{eq:scm} need not average
to zero over $N$ units; we draw a large population in each sandbox to keep the
discrepancy below the reporting precision of every effect measure.

\paragraph{Free instances.}
$p$ is drawn uniformly from a difficulty-tier range and edges are placed by a
fixed-order Erd\H{o}s--R\'enyi rule: node $i$ receives
$\mathrm{Binomial}(i{-}1, \pi)$ parents sampled without replacement from
$X_1,\dots,X_{i-1}$, with $\pi$ set so the expected edge count matches a target
that scales with $p$, from $\log p$ at the sparsest tier to $p(p{-}1)/4$ at the
densest. The fixed topological order rules out cycles by construction. One
endogenous node is then designated $Y$, and the whole graph is re-drawn until
that node's in-degree and number of upstream colliders both fall inside the
study's target range. These are the two structural knobs
Section~\ref{sec:design} varies independently.

\paragraph{Canonical instances.}
The five shapes of Figure~\ref{fig:topologies} are placed as an explicit edge
list
parameterized by one integer level: fork, for example, is $D \to Y$ plus one
$Z_i \to Y$, $Z_i \to W_i$ pair per level, and similarly for the other four.
Sweeping one shape's size while every other identification descriptor stays
fixed is what a free random graph cannot guarantee on demand.

\paragraph{Mechanism family.}
The engine also supports non-linear and tabular mechanisms. Every study reported
here uses the linear family, which is what keeps the effects closed-form and
supplies the $\sigma^{2}$ that the budget rule of Section~\ref{sec:budget}
depends on.

\section{Runner and termination}
\label{app:pipeline}

\paragraph{Provider abstraction.}
Every provider is driven through one interface: the runner emits a
provider-agnostic list of messages and tool specifications and receives back a
normalized response carrying reasoning text, at most one tool call, a stop
reason and token usage. Provider-specific payload that must round-trip verbatim to keep a
multi-turn tool-call chain valid, such as reasoning content some providers
require replayed on the next turn or a thought signature others attach to a
tool call, is carried as opaque data the runner never inspects. Responses
carrying several tool calls in one turn are normalized to at most one. This
uniform contract, backed by two adapter implementations, is what lets one
harness run every model in Section~\ref{sec:experiments} without a per-model
branch in the loop.

\paragraph{Termination.}
An episode ends one of four ways: the agent submits; the budget reaches zero;
the iteration cap is reached; or the adapter fails. Exhausting the budget or the iteration cap does not end the episode outright.
The agent is given one further turn on which observation and experiment are
unavailable but submission is still possible. An agent that never submits is scored as having submitted
the empty set rather than excluded from the corpus.

\section{Instance descriptors}
\label{app:descriptors}

Twenty-five descriptors are recorded for every instance, grouped by the obstacle
they belong to. \emph{Identification} (seven): topology, confounders, colliders,
mediation chains, longest chain, non-intervenable variables, intervenable-parent
fraction. \emph{Estimation} (five): smallest true effect, largest distractor
effect, effect spread, signal confusion, $n^{\ast}$. \emph{Design cost} (four):
$B_{\mathrm{test}}$, two sufficiency ratios, the budget multiplier. A further
five record the scale of the search and four the provenance of the instance, so
that any instance in the corpus can be regenerated exactly.

\section{The forty measures}
\label{app:metrics}

All forty are written by one code path for every fully scored episode. Nine
cover structure recovery, of which \emph{exact recovery} is reported throughout;
two cover effect estimation; six attribute each wrong edge to the evidentiary
state the agent had reached for that candidate --- never tested, tested without
control, underpowered, unresolved at submission, or resolved and misread; eleven
describe what the agent ran, of which \emph{controlled experiments per episode}
is reported throughout; two cover sample sizing; five cover repetition and
stability; and five cover resource use. Four of the forty are inert on this
corpus, two constant and two duplicating another measure, leaving thirty-six
live.

Attribution uses three fixed constants: a minimum detectable effect of $0.5$, a
pooled $|t|$ threshold of $2.5$, and an effect gate of $0.25$. These follow from
the generator's coefficient range rather than from any episode. Two scope
conditions apply. The ladder is read from interventional contrasts only, so a
candidate an agent resolved by regression on observational rows counts as never
tested, which is why the toolbox arms of Section~\ref{sec:exp-levers} are
reported on outcome measures alongside it. And because several contrasts on one
candidate are pooled by inverse variance, a later contrast that disagrees with
an earlier one can move a candidate between the upper states; only the
never-tested boundary is monotone.

\paragraph{The decision rule.}
The attribution charges agents with wrong answers to the first step they
fail, reading only their own \texttt{experiment} calls and pooling them by
inverse variance.
\begin{enumerate}\itemsep0pt\parskip0pt
\item No experiment touched it.
\item Its calls held every other candidate fixed, but the pooled estimate
      clears neither of the thresholds above, so it neither establishes the
      effect nor rules it out.
\item The only significant result comes from calls that did not hold the
      others fixed, which cannot separate a parent from an ancestor, and the
      candidate was reported as a parent.
\item The pooled estimate clears one of them and the submitted answer
      disagrees with it.
\item Anything else still unresolved when the agent submitted.
\end{enumerate}

\paragraph{What the observational rows would change.}
The rule above ignores the rows the agent bought with \texttt{observe}, so we
measured how much of the attribution depends on that. For every episode that
still has its raw rows, we ran the adjustment ourselves on the rows that agent
had bought, and credited the candidate at the first level whenever our
regression got it right. The agent never ran that regression, so this credits
it with work it did not do, and the second column is an upper bound on what the
exclusion is worth rather than a corrected attribution.

\begin{center}
\small
\begin{tabular}{lrr}
\toprule
\textbf{Level} & \textbf{Interventional only} & \textbf{Either route} \\
\midrule
no identifying evidence obtained         & $20.5\%$ & $1.0\%$ \\
candidate never touched                  & $62.3\%$ & $19.2\%$ \\
obtained but underpowered                & $11.2\%$ & $11.2\%$ \\
evidence in hand, no answer drawn from it & $2.4\%$  & $65.0\%$ \\
submitted while still unresolved         & $3.6\%$  & $3.6\%$ \\
\midrule
first three levels combined              & $94.0\%$ & $31.3\%$ \\
\bottomrule
\end{tabular}
\end{center}

Both columns cover the same 3,643 errors, and read $94.0\%$ rather than
the body's $94.4\%$ because episodes without raw rows are left out. The
conflict-resolution case the paper defines stays at $2.4\%$, and the other
$62.6\%$ of the fourth row is the credit described above. The evidence was
usually bought already: 2,298 of these 2,862 episodes held enough rows, and on
them the adjustment settles 2,780 of the agent's 2,950 wrong candidates.

\paragraph{The attribution across models.}
The attribution is computed on the whole corpus, but the ablations that
motivate it were run on \texttt{claude-sonnet-4-6}, so the table below reads
the same ladder on every model carrying at least twenty attributed errors.
Three of the four carry few errors, so the counts are given alongside the
shares.

\begin{center}
\small
\begin{tabular}{lrr}
\toprule
\textbf{Model} & \textbf{Attributed errors} & \textbf{First three levels} \\
\midrule
\texttt{claude-sonnet-5}   & 30    & $96.7\%$ \\
\texttt{claude-sonnet-4-6} & 2,876 & $94.2\%$ \\
\texttt{qwen3.7-plus}      & 51    & $92.2\%$ \\
\texttt{gpt-5.6-terra}     & 134   & $85.8\%$ \\
\bottomrule
\end{tabular}
\end{center}

A two-parameter logistic item-response fit \citep{embretson2000irt} over 13
models and 41 instances converges with ability correlating to raw score at
$r = 0.96$; the 8 instances below a discrimination of $0.5$ are reported rather
than dropped.

\section{Sampler robustness}
\label{app:sampler}
Repairing the sampler moves nothing, and we bound how much it could have
moved. An earlier version of the environment drew population rows with
replacement, which shrinks standard errors as $1/\sqrt{n}$ without adding
information. We reran an entire $4\times4$ design under both samplers with
everything else held fixed, in the regime where the artifact should bite
hardest: no per-call cap and budgets to 125 times, where a single call can draw
a large share of the population. Over 480 matched pairs the difference in exact
recovery is $-0.006$, 95\% bootstrap CI $[-0.033, +0.021]$
\citep{efron1979bootstrap}, with 433 of 480 episodes agreeing exactly, and no
trend across the budget axis along which exposure rises. We therefore bound
the artifact at four percentage points on the budget and tool conclusions,
where exposure is highest; the remaining studies run under a per-call cap and
are less exposed still.

\section{Distinction from Relevant Concurrent works}
\label{app:concurrent}
Our work is a focused diagnosis-and-intervention study of LLM causal discovery under finite budgets for observational and experimental data acquisition. By restricting the identification task to an outcome variable’s direct causal parents and their treatment effects, we support fine-grained diagnosis of failures in experimental contrast design, candidate coverage, sample allocation, and evidence resolution. We pair these diagnostics with targeted interventions to investigate which forms of assistance improve task performance and how they change the way agents conduct sequential experiments.

Relative to the four compared benchmarks, including the concurrent works ACDB, CausalArena, and ACED-Bench, our study characterizes two specific aspects of agent behavior that are not directly examined in these four benchmarks. First, through ablation studies that provide agents with data from controlled versus uncontrolled experiments for the same causal systems and identification tasks, we test whether agents can correctly navigate between experimental evidence that isolates direct treatment effects from experiments that only show total treatment effects (i.e., those that also include indirect treatment effects through mediated paths). Second, through ablation studies on whether to impose caps on sample sizes, we test whether the agents can balance budget constraints with statistical-power requirements and examine agents’ sensitivity to the identification requirements of the task as well as the trade-off between statistical precision and sampling costs.

Our intervention studies distinguish the different ways assistance can improve causal discovery. Although related interventions appeared in other works, i.e., ACDB examines statistical-tool access, ACED-Bench studies memory and resource protocols, and CausaLab tests evidence-consistency verification, our work makes a distinctive contribution by connecting targeted assistance to specific failure mode mitigation in sequential data acquisition and experimental design. For example, increasing budgets can enhance experiment coverage for potential causal ancestors, but it does not reliably elicit controlled experimentation in our evaluated setting. For LLM agents that lack adequate experiment controls, we instead propose mitigating this issue with statistical tools that offer an alternative route to causal inference. In contrast, sample size guidance and explicit prompt engineering are designed to guide experimental procedures and can change how agents allocate samples and conduct controlled versus uncontrolled experiments. These interventions increase controlled experimentation, while final recovery varies across models and settings. Our findings provide both guidance on how to pair assistance with diagnosed agent-specific performance bottlenecks and how to assess the effectiveness of proposed assistance through both final task accuracy and changes in the agent's sequential experiment design.

\section{Nonlinear Experimental Results}
\label{app:nonlinear-results}
We report the nonlinear budget, guidance, model-comparison, and tool ablations below. Curve error denotes the median normalized $L_1$ error in Eq.~\ref{eq:nonlinear_curve_error}; lower is better. The nonlinear evaluation comprises 2,925 completed episodes across 13 models.

\begin{table}[htbp]
\centering\small
\caption{Nonlinear budget ablation. Controlled experiments and curve error are medians; paired ratios are relative to $1\times$.}
\label{tab:nl-budget}
\resizebox{\linewidth}{!}{\begin{tabular}{@{}lrrrr@{}}
\toprule
Budget & Controlled exps. & Controlled rate & Curve error & Paired vs. $1\times$ \\
\midrule
$1\times$ & 4 & 0.53 & 0.1866 & -- \\
$5\times$ & 5 & 0.55 & 0.0981 & 0.58 \\
$25\times$ & 8 & 0.64 & 0.0743 & 0.38 \\
$125\times$ & 14 & 0.70 & 0.0534 & 0.28 \\
\bottomrule
\end{tabular}}
\end{table}

The same-instance scripted solver provides a reference for nonlinear task solvability at each budget level (Table~\ref{tab:nl-scripted-solver}). It substantially outperforms the agents in both exact parent-set recovery and curve estimation across all four budgets.

\begin{table}[htbp]
\centering\small
\caption{Nonlinear scripted solver versus agents on the same instances across budget levels. Curve error is lower-is-better.}
\label{tab:nl-scripted-solver}
\resizebox{\linewidth}{!}{\begin{tabular}{@{}lrrrr@{}}
\toprule
Budget & Solver exact & Solver curve error & Agent exact & Agent curve error \\
\midrule
$1\times$ & 82.0\% & 0.0449 & 9.7\% & 0.1866 \\
$5\times$ & 94.0\% & 0.0317 & 37.5\% & 0.0981 \\
$25\times$ & 96.7\% & 0.0295 & 67.3\% & 0.0743 \\
$125\times$ & 96.7\% & 0.0295 & 72.9\% & 0.0534 \\
\bottomrule
\end{tabular}}
\end{table}

\subsection{Prompt arms}
\label{app:prompt-arms}

The prompt ablation has four arms. V0 is the baseline system prompt the agent
receives in every other study, unchanged. V1, V2 and V3 append one or more
guidance modules to that same baseline, verbatim and in the order listed below;
nothing is removed and no tool, budget or interface setting differs across the
four arms.

\begin{center}
\small
\begin{tabular}{llll}
\hline
\textbf{Arm} & \textbf{Guidance} & \textbf{Modules appended} & \textbf{Linear arm} \\
\hline
V0 & baseline, none & ---                  & V0 \\
V1 & objective                 & T1, T2, T3, T5       & V1 \\
V2 & experimental strategy     & S1                   & V2, rewritten \\
V3 & objective + strategy      & T1, T2, T3, T5, S1   & none \\
\hline
\end{tabular}
\end{center}

\paragraph{The objective modules (V1, V3).}

{\small
\begin{verbatim}
T1  Objective: submit only the DIRECT visible parents of the outcome.
    Exclude indirect causes, ancestors, descendants, mediators and any
    other non-parent.

T2  You are scored on parent recall, parent precision, the accuracy of
    each per-unit effect, and the accuracy of each dose-response CURVE.

T3  Submission: `edges` are [source, outcome] pairs and `curves` is a
    same-length list of closed-form expressions in `x`. The per-unit
    effect of an edge is the least-squares slope of its curve against
    the dose, so a two-dose contrast (Y_b - Y_a)/(b - a) estimates it
    only if the curve is a straight line.

T5  Stopping: stop when a further experiment is unlikely to change the
    parent set or materially improve a submitted curve.
\end{verbatim}
}

\paragraph{The strategy module (V2, V3).}

{\small
\begin{verbatim}
S1  Strategy. 1) Take an observational sample. 2) Screen each candidate
    with a CONTROLLED contrast: intervene on it while pinning every
    other candidate, so a mediated path cannot carry the effect. 3) For
    a surviving parent, identify the dose-response shape by sweeping at
    least four distinct doses across the admissible range in one or more
    experiments, pinning the other parents, then comparing the per-dose
    means against each candidate shape at several scales. 4) Spend
    remaining budget where it most reduces uncertainty per unit cost.
    5) Stop once the outcome's neighbourhood and its shape are resolved.
\end{verbatim}
}

\paragraph{Provenance, and where the arms depart from the linear study.}
T1 and T5 restate the linear study's modules of the same name. T2 and T3 are
adjusted because the estimand changed: T2 adds the dose-response curve to the
list of scored quantities, and T3 replaces the linear definition of the
per-unit effect as the two-point difference quotient
$(Y_b-Y_a)/(b-a)$ with the least-squares slope of the submitted curve, which is
the quantity the scorer actually uses and which the difference quotient
recovers only when the curve is a straight line. S1 is not a restatement. The
linear strategy module ran to roughly fifty lines in two phases --- PC-Stable
screening of a CPDAG, then greedy interventional refinement --- and the present
S1 compresses it to five numbered steps, drops the CPDAG construction
entirely, and adds step 3, which has no linear counterpart because the linear
task has no shape to identify. S1 is therefore a more explicit instruction to
run the controlled contrast than its linear predecessor, and the resulting
change in controlled-experiment counts should not be read as a property of the
models alone. V3 has no linear counterpart at all and is reported as a new arm.

The prompt-ablation results appear in Table~\ref{tab:nl-prompts}, with model-level comparisons in Tables~\ref{tab:nl-weak-guidance} and~\ref{tab:nl-model-comparison}.

\begin{table}[htbp]
\centering\small
\caption{Nonlinear prompt ablation. Controlled experiments and curve error are medians; fully controlled episodes are counts.}
\label{tab:nl-prompts}
\resizebox{\linewidth}{!}{\begin{tabular}{@{}lrrrrr@{}}
\toprule
Prompt & Controlled exps. & Controlled rate & Fully controlled & Curve error & Paired vs. V0 \\
\midrule
V0 baseline & 4 & 0.54 & 32/195 & 0.1017 & -- \\
V1 objective & 5 & 0.59 & 48/194 & 0.1089 & 1.02 \\
V2 strategy & 9 & 0.97 & 178/195 & 0.0779 & 0.80 \\
V3 combined & 9 & 0.95 & 167/195 & 0.0784 & 0.89 \\
\bottomrule
\end{tabular}}
\end{table}

\begin{table}[htbp]
\centering\small
\caption{Effects of strategy guidance on models that rarely run controlled experiments without guidance. Values are medians; V0 denotes baseline and V2 denotes strategy guidance.}
\label{tab:nl-weak-guidance}
\resizebox{\linewidth}{!}{\begin{tabular}{@{}lrr@{}}
\toprule
Model & Controlled exps. V0 $\to$ V2 & Curve error V0 $\to$ V2 \\
\midrule
claude-sonnet-4-6 & $0\to9$ & $0.5029\to0.2457$ \\
claude-sonnet-4-6-thinking & $0\to8$ & $0.4452\to0.3652$ \\
qwen3.7-plus & $1\to7$ & $0.5981\to0.1618$ \\
All three pooled & $0\to8$ & $0.5077\to0.2457$ \\
\bottomrule
\end{tabular}}
\end{table}

\begin{table}[htbp]
\centering\small
\caption{Unprompted stronger models (V0) versus guided weaker models (V2) on nonlinear tasks. Controlled experiments and curve error are medians.}
\label{tab:nl-model-comparison}
\resizebox{\linewidth}{!}{\begin{tabular}{@{}llrr@{}}
\toprule
Model & Condition & Controlled exps. & Curve error \\
\midrule
Gpt 6 astra & V0 & 9 & 0.0512 \\
Gemini 3.1 Pro Preview & V0 & 12 & 0.0549 \\
Qwen3.8 max & V0 & 10 & 0.0695 \\
Claude fable 5.1 & V0 & 8 & 0.0735 \\
Claude sonnet 5 & V0 & 8 & 0.0825 \\
Gpt-5.6 Sol & V0 & 6 & 0.0902 \\
Gemini 3.8 Flash & V0 & 5 & 0.1047 \\
DeepSeek V4 Pro & V0 & 7 & 0.1195 \\
\midrule
Qwen3.7 plus & V2 & 7 & 0.1618 \\
Claude Sonnet 4.6 & V2 & 9 & 0.2457 \\
Claude Sonnet 4.6, thinking & V2 & 8 & 0.3652 \\
\bottomrule
\end{tabular}}
\end{table}

\begin{table}[!htbp]
\centering\small
\caption{Nonlinear tool ablation. Curve error is the median; geometric mean and paired ratios are as reported for each arm.}
\label{tab:nl-tools}
\resizebox{\linewidth}{!}{\begin{tabular}{@{}lrrrr@{}}
\toprule
Arm & Optional tools & Curve error & Geometric mean & Paired vs. A0 \\
\midrule
A0 core only & 0 & 0.0941 & 0.1007 & 1.00 \\
A4 shape fitting & 1 & 0.0761 & 0.0660 & 0.66 \\
A5 curve toolkit & 2 & 0.0814 & 0.0877 & 0.87 \\
A2 basic stats & 3 & 0.1063 & 0.0927 & 0.92 \\
A3 causal structure & 4 & 0.1107 & 0.1352 & 1.34 \\
A1 all tools & 16 & 0.1017 & 0.0886 & 0.88 \\
\bottomrule
\end{tabular}}
\end{table}

\begin{table}[!htbp]
\centering\small
\caption{Use of \texttt{regress\_basis} when available alone (A4) or within the sixteen-tool suite (A1), over 390 episodes per arm.}
\label{tab:nl-tool-usage}
\resizebox{\linewidth}{!}{\begin{tabular}{@{}lrrrr@{}}
\toprule
Arm & Optional tools & Total calls & Median calls/episode & Episodes using it \\
\midrule
A4 shape fitting & 1 & 2,135 & 4.5 & 377/390 \\
A1 all tools & 16 & 1,549 & 3.7 & 374/390 \\
\bottomrule
\end{tabular}}
\end{table}

\end{document}